\pdfoutput=1 
\documentclass[acmsmall]{acmart}
\usepackage{multirow}
\usepackage[ruled,lined,boxed,commentsnumbered]{algorithm2e}
\usepackage{fancyhdr}
\renewcommand\footnotetextcopyrightpermission[1]{}

\AtBeginDocument{%
  \providecommand\BibTeX{{%
    \normalfont B\kern-0.5em{\scshape i\kern-0.25em b}\kern-0.8em\TeX}}}

\begin{document}

\title{Vis-TOP: Visual Transformer Overlay Processor}

\author{Wei Hu}
\authornote{all the authors contributed equally to this research.}

\email{huwei@wust.edu.cn}
\affiliation{
  \institution{Wuhan University of Science and Technology}
  \streetaddress{Huangjiahu West Road}
  \city{Wuhan}
  \state{Hubei}
  \country{China}
  \postcode{430065}
}
\author{Dian Xu}
\email{dianhsu@wust.edu.cn}
\authornotemark[1]
\affiliation{
  \institution{Wuhan University of Science and Technology}
  \streetaddress{Huangjiahu West Road}
  \city{Wuhan}
  \state{Hubei}
  \country{China}
  \postcode{430065}
}
\author{Zimeng Fan}
\authornotemark[1]
\email{fanzimeng@wust.edu.cn}
\affiliation{
  \institution{Wuhan University of Science and Technology}
  \streetaddress{Huangjiahu West Road}
  \city{Wuhan}
  \state{Hubei}
  \country{China}
  \postcode{430065}
}
\author{Fang Liu}
\email{liufangfang@whu.edu.cn}
\authornotemark[1]
\affiliation{
  \institution{Wuhan University}
  \streetaddress{Bayi Road}
  \city{Wuhan}
  \state{Hubei}
  \country{China}
  \postcode{430072}
}

\author{Yanxiang He}
\email{yxhe@whu.edu.cn}
\authornotemark[1]
\affiliation{
  \institution{Wuhan University}
  \streetaddress{Bayi Road}
  \city{Wuhan}
  \state{Hubei}
  \country{China}
  \postcode{430072}
}

\begin{abstract}
    In recent years, Transformer \cite{Vaswani2017} has achieved good results in Natural Language Processing (NLP) and has also started to expand into Computer Vision (CV). Excellent models such as the Vision Transformer \cite{Dosovitskiy2020} and Swin Transformer \cite{liu2021swin} have emerged. At the same time, the platform for Transformer models was extended to embedded devices to meet some resource-sensitive application scenarios. However, due to the large number of parameters, the complex computational flow and the many different structural variants of Transformer models, there are a number of issues that need to be addressed in its hardware design. This is both an opportunity and a challenge. We propose Vis-TOP (Visual Transformer Overlay Processor), an overlay processor for various visual Transformer models. It differs from coarse-grained overlay processors such as CPU, GPU, NPE, and from fine-grained customized designs for a specific model. Vis-TOP summarizes the characteristics of all visual Transformer models and implements a three-layer and two-level transformation structure that allows the model to be switched or changed freely without changing the hardware architecture. At the same time, the corresponding instruction bundle and hardware architecture are designed in three-layer and two-level transformation structure. After quantization of Swin Transformer tiny model using 8-bit fixed points (fix\_8), we implemented an overlay processor on the ZCU102. Compared to GPU, the TOP throughput is 1.5x higher. Compared to the existing Transformer accelerators, our throughput per DSP is between 2.2x and 11.7x higher than others. In a word, the approach in this paper meets the requirements of real-time AI in terms of both resource consumption and inference speed. Vis-TOP provides a cost-effective and power-effective solution based on reconfigurable devices for computer vision at the edge.
\end{abstract}

\maketitle

\section{Introduction}

With the concept of artificial intelligence becoming popular throughout the world, deep neural networks have been widely used in a variety of fields, from RNN, LSTM in Natural Language Processing and CNNs in Computer Vision, all of which have achieved state of the art results and become the primary choice in many fields. Transformer \cite{Vaswani2017}, which simulates human’s attention through a unique attention mechanism, has attracted many researchers in the field of natural language processing, and a number of excellent models have emerged \cite{devlin2018bert,sun2020mobilebert,jiao2019tinybert,Wu2020,lan2019albert,liu2019roberta}. In recent years, the Transformer has started to expand into computer vision. Vision Transformer \cite{Dosovitskiy2020} was the first to propose a network model that completely replaces the CNN structure with Transformer, and has achieved very good results. Thanks to that, More and more visual Transformer models appear \cite{pan2021scalable,2021,touvron2021training}.

On the other hand, as research about network model deepens, researchers are beginning to focus not only on model accuracy, but also on metrics that are critical in real-world application scenarios such as resource utilization, operation efficiency, resource consumption and latency. Although GPUs can often deploy network models by leveraging the multiple cores they provide, their power consumption is high. Moreover, its architecture is universal, which results in a lack of customization support for each algorithm. Thanks to customizable feature, reconfigurable devices become another optional direction. It offers better operational efficiency and lower power consumption than universal processors such as GPUs. The solution based on reconfigurable devices is the following: 1) analyze the structure of neural networks and core algorithms. 2) design customized hardware architectures. 3)implement data and computational flows with strong parallelism and pipelining based on the entire model flow. 4) optimize and iterate on original solutions based on reports. There are many great works for CNN \cite{Gysel2018,DBLP:conf/fpga/ZhangLSGXC15,Yang2018}, RNN or LSTM \cite{bank2019polar,liao2017fpga,guan2017fpga}. But there are relatively few works for Transformer models, such as \cite{Khan2021,Li2020,Park2020}.

In current research, the evolution of algorithms is mainly supported by computation power, which consists of two aspects of computation support, namely high computation power with GPU stacking and acceleration by designing a specific hardware architecture or co-processor for a specific algorithm. The problem with GPU stacking is that universal architectures are difficult to adapt to the characteristics of different algorithms, while customized hardware architecture can only support a specific algorithm and is not scalable. This paper proposes a design for visual Transformer overlay processor based on the co-evolution of software and hardware, and the main contributions of this paper can be summarized as follows:
\begin{itemize}
    \item This paper proposes a model transformation and computation architecture. It is a complete architecture consisting of a model layer, a container layer and a component layer, with a two-level transformation relationship between the model-container and container-component layers. In this paper, an instruction bundle is also designed to support the two-level transformation relationship. The three-layer architecture and the two-level transformation relationship build a transparent mapping from the core of algorithm to the hardware, allowing the deployment of neural network while shielding the underlying hardware and enabling customization for different neural networks. This creates synergy between the neural network and the hardware structure, which provide great flexibility and improve computational efficiency.
    \item We propose and design a hardware architecture that is compatible with a three-layer and two-level transformation structure. The architecture starts by slicing the neural network model from a hardware perspective, providing basic support to the upper layers in the form of "universal modularity + reconfigurability" at the lowest level. Instead of considering the characteristic and logical significance of neural network model, it provides basic support to the upper layers in the form of fixed module with reconfigurable batch and variable module with reconfigurable structure. In this case, these modules are derived from common modules after slicing the neural network. The upper layer can configure two types of modules by means of instruction bundles and perform parametric configuration according to the characteristics of neural network and the algorithm themselves. Due to this, the upper layers can determine the order of execution of these modules in order to implement the overall flow of the network model more quickly.
    \item By analyzing the data storage, data selection and data computation of these visual Transformers, the paper designs fixed modules and variable modules that reflect the characteristics of different visual Transformers. Swin Transformer is used as a case study for testing. After quantifying tiny version of \cite{liu2021swin} using 8-bit fixed points (fix\_8), an overlay processor is implemented on the ZCU102 and the resource consumption and inference time of each module is evaluated. The final hardware experimental results show that Vis-TOP has 1.5x the throughput of GPUs compared to same implementation on the GPU. Compared to the existing Transformer class accelerators based on FPGA, Vis-TOP's Throughput per DSP also is 2.2x to 11.7x.
\end{itemize}

The rest of this paper is organized as follows. Section II provides a brief introduction to the Transformer as well as the visual Transformer. Section III summarize and present current research work on Customized accelerators and overlay processors. In Sections IV and V, we introduce the three-layer and two-level transformation structure in hardware and software co-design proposed in this paper, as well as a detailed description of the proposed hardware architecture. Section V Section 5 will give quantification scheme and experimental results. A summary of the research will be presented in Section VI.

\section{Background}
\subsection{Visual Transformer}

\begin{figure}[ht]
    \centering
    \includegraphics[width=\linewidth]{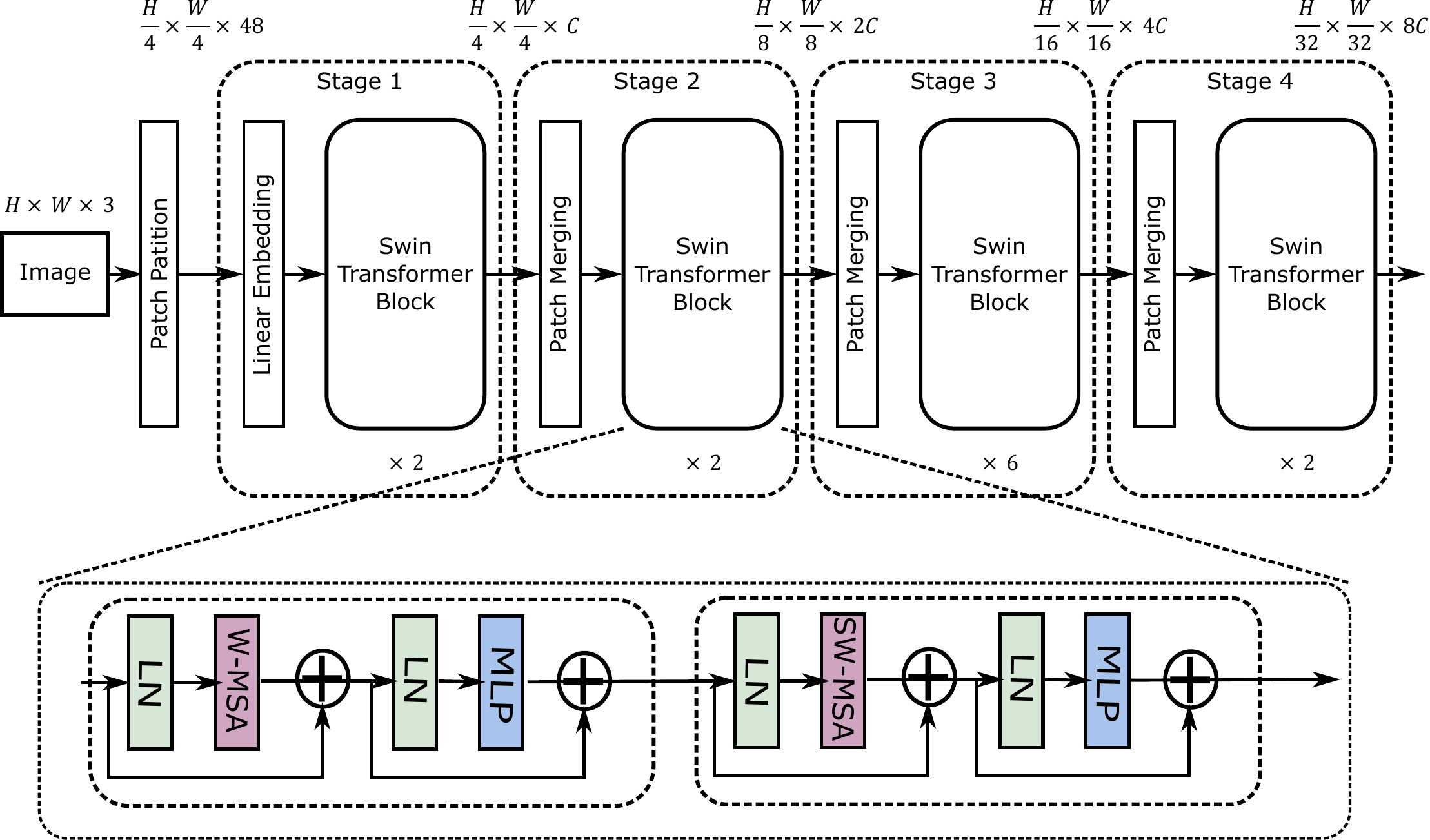}
    \caption{The model architecture of Swin Transformer}
    \Description{The model architecture of Swin Transformer}
    \label{img:swintransformer}
\end{figure}

Since the introduction of \cite{Dosovitskiy2020}, various Transformer models for machine vision start to appear in large numbers, and Swin Transformer \cite{liu2021swin} is one of the best models. We use it as an example to introduce the visual Transformer. As shown in Fig.~\ref{img:swintransformer}, It refers to the structure of CNN and divides the whole model into 4 different stages. The input feature map is down sampled at each stage (the height and width of the image is reduced and the number of channels is increased), and only the local area is modelled in the block of each stage instead of the global area.
Window based multi-head attention(W-MSA) only models the interrelationships in a window (7x7 in this paper), which helps to reduce the overall computational complexity, similar to the idea of moving from full connectivity to convolution. In addition to the similarity of ideas, W-MSA also has the same kernel size and stride parameters as the convolution operation, but the difference is that W-MSA has the same kernel size and stride in a stage, which leads to a problem that the receptive field of all W-MSAs in each stage are same. Both CNN and MSA do not have this problem: CNN has a larger kernel size than stride, so its receptive field increases gradually as the convolution stacks up; MSA has a global receptive field because the kernel size completely covers the feature map.
To alleviate this problem, the authors propose SW-MSA (Shifted Window based Multi-head Self-attention). The only difference between SW-MSA and W-MSA is that the window is shifted to artificially change receptive field. This problem is solved by using SW-MSA and W-WSA alternately in the same block.

\section{Related work}
This chapter will summarize and introduce the related works in 2 aspects: customized accelerator and overlay processor. This paper will first introduce customized hardware accelerator for Transformer models on embedded devices (ASICs and FPGAs). Then, this paper will introduce the current works related to overlay processor. Although Transformer applications and research in various fields have been abundant,  the overlay processor for Transformer models is even only NPE \cite{Khan2021}, so we will compare some overlay processor designs for CNNs.

\subsection{Customized accelerators}

A$^3$ \cite{Ham2020a} devised a customized hardware accelerator to accelerate the single-round attention mechanism. It approximates the actual value by obtaining the max and min elements of the Fixed sort as candidates. OPTIMUS \cite{Park2020} proposes skipped redundant computations for redundant operations in decoder and Set-Associative RCSC (SA-RCSC) format to increase MAC utilization by enabling multiple PEs to process a single matrix row. A hardware accelerator \cite{lu2020hardware} is proposed to accelerate MHA ResBlock and FFN ResBlock in the Transformer. For non-linear functions such as Softmax and LayerNorm, it analyses and designs detailed computational flows and hardware structures. \cite{liu2021hardware} investigates the acceleration of BERT on FPGAs. To reduce the memory footprint, the authors fully quantize all parameters of BERT, including weights, activations, scale factors, SoftMax, layer normalization and other intermediate results. FTRANS \cite{Li2020} is a FPGA accelerator also for BERT. It proposes an enhanced BCM (Block-Circulant Matrix) based method for feed-forward operations. The original weight matrix is replaced by one or more circular matrix blocks to reduce weight storage. It proposes an architecture that supports model compression techniques and develops an optimization algorithm for design automation and exploration of parallelism.

\subsection{Overlay processor}
DLA \cite{abdelfattah2018dla} implements a domain specific approach to overlays via a lightweight very-long instruction word network, where the authors connect the core functions of neural networks with parameterizable interconnects to implement different model configurations. OPU \cite{Yu2020}proposes an FPGA-based overlay processor system for general convolutional neural networks. The authors propose a decomposition of the convolution operation into convolution-flattening (CF) and inner-product (IP). Thanks to this decomposition, the processor can implement different convolutional operations (such as different convolutional kernel sizes) in the same set of modules. At the same time the authors define fine-grained OPU instructions and automatic compiler to explore parallelization. In contrast, Light-OPU \cite{Yu2020a} and Uni-OPU \cite{Yu2020b} extend architecture based on OPU by adding light-weight CNNs and transposed CNNs respectively. NPE\cite{Khan2021} is the first overlay processor dedicated to natural language processing (NLP) on a reconfigurable device. The authors propose a system that can handle any non-linear function by means of piecewise polynomial approximation along with simple vector operations. The system can be adapted to all types of models, leaving a huge design space.

\section{Three-layer and two-level transformation structure}

\begin{figure}[ht]
    \centering
    \includegraphics[width=\linewidth]{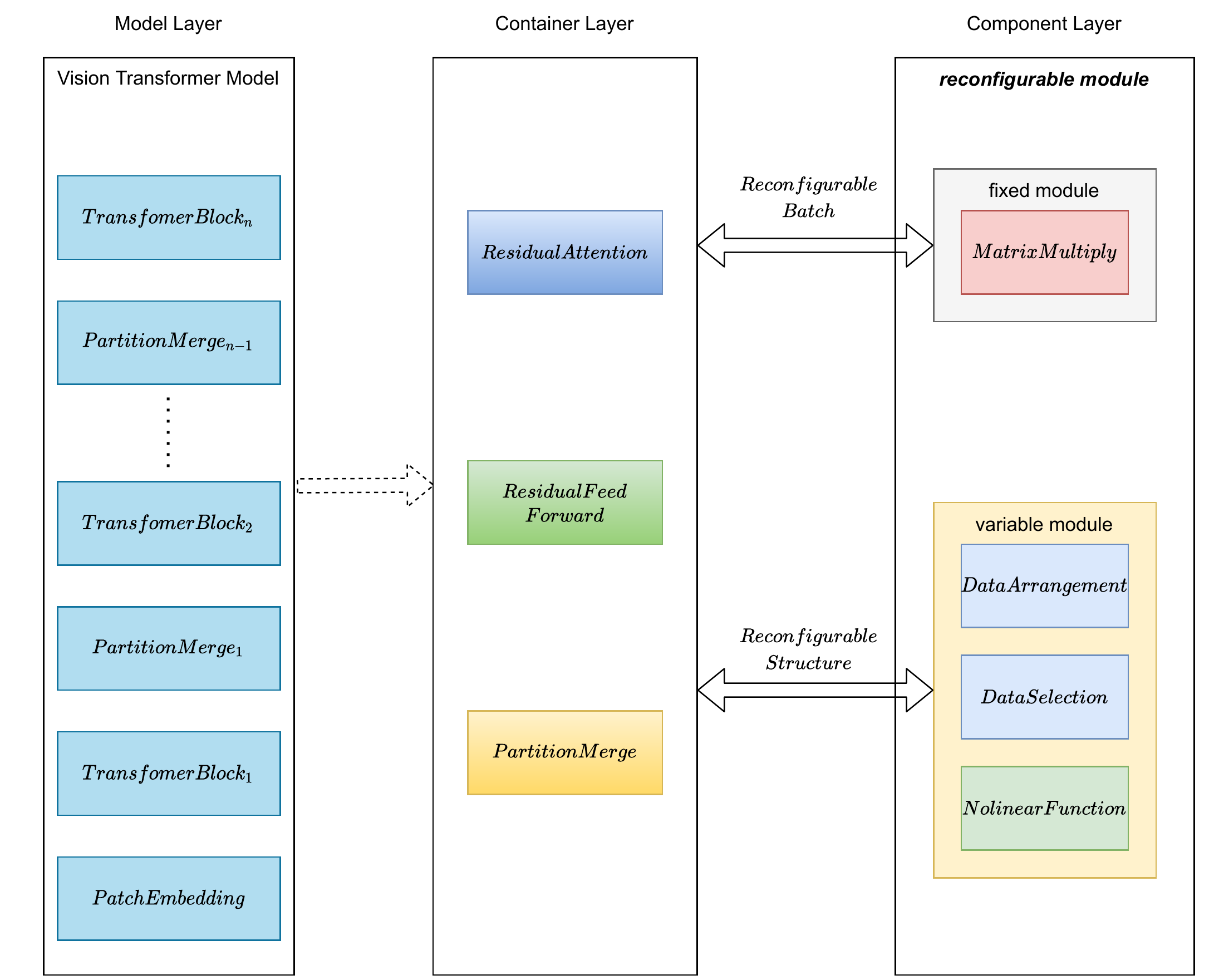}
    \caption{Three-layer and two-level transformation structure}
    \Description{Three-layer and two-level transformation structure}
    \label{img:Three-layer}
\end{figure}

Taking inspiration from the traditional design steps of overlay accelerators, is it possible that we can deploy neural networks while shielding the underlying hardware design, which significantly reduces the development cycle and requirements? Based on the idea of co-optimization of hardware and software, this paper designs an architecture consisting of three layers, namely the model layer, the container layer and the component layer, as well as a 2-level transformation relationship based on the three levels.The overall structure is shown in Fig.~\ref{img:Three-layer}.

\textbf{Model}: In this layer we decompose the target network model into modules, which are usually divided by the functions implemented at the software level.  For example, ConVit \cite{2021} divides itself into GPSA, FFN and SA modules; Swin Transformer \cite{liu2021swin} into partition merge, residual feed-forward, residual attention modules; Scalable Visual Transformer \cite{pan2021scalable} is further divided into Linear Projection, Transformer Block, max pool and other modules. Each module has a number of hyper-parameters to control the properties of these module, for example, the position-wise feed-forward dimension, which is present in all Transformer models, the hidden layer dimension, the word embedding dimension and the number of layers in the linear layer, etc. These attributes greatly affect the resource consumption and computation time of the corresponding modules.

\textbf{Container}: Traditional research progress is actually achieved by "algorithm" and "computing power" respectively. This is a system construction method of separating software and hardware, which makes it difficult to intuitively embody the concept of hardware-software co-design. This also makes it difficult to achieve optimum operational efficiency because of separating the hard and soft aspects of the overall design. Therefore, the container layer is added to better reflect the software-to-hardware and hardware-to-software transformation and compatibility process.

In this layer we propose the instruction bundle, which is a logical structure. It is essentially a transition layer between software and hardware, and mainly reflects our idea of hardware-software transformation from Model to Component. The transformation relationships for software-level structures are above the Instruction Bundle in this layer, while the transformation relationships for hardware-level components are below the Instruction Bundle in this layer. The Instruction Bundle provides both public and private extensions to meet the algorithmic needs of some particular model. The public instruction bundle provides an interface to computation modules for general-purpose operations; the private instruction bundle provides an interface to computational modules for algorithmic characteristics.

\textbf{Component}: This layer consists of all the fixed dimensional modules solidified into hardware logic. The data dimensions to be processed are defined at design time and have a network model independent nature. For a specific model, it can be divided into two components, i.e., fixed module and variable module. The fixed module is a module for general-purpose operations. These modules take a fixed amount of time each time. It can achieve different scales of calculations by setting different batch values. The variable  module is a module for algorithmic calculation features. It consists of nonlinear function modules, data selection module and data arrangement module. The run times for these modules are related to the size of input data or parameters. The nonlinear function module needs to be customized for specific nonlinear functions, while the data selection module and data arrangement module need to be designed according to the size and shape of the data blocks for subsequent modules.

\section{Hardware architecture}

\subsection{Overall architecture}
\begin{figure}[ht]
    \centering
    \includegraphics[width=\linewidth]{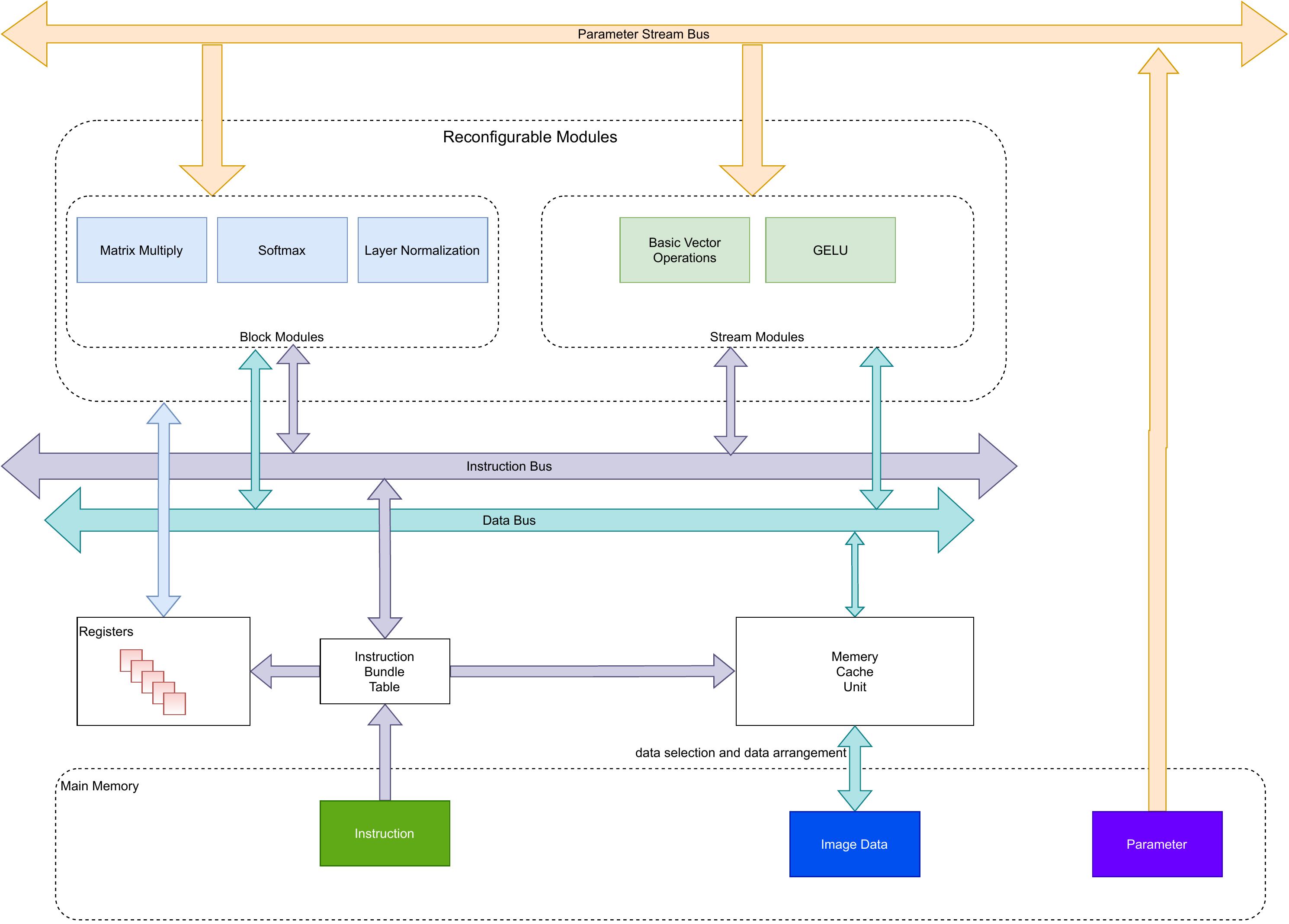}
    \caption{The structure of our hardware design}
    \Description{The structure of our hardware design}
    \label{img:hardware structure}
\end{figure}

This hardware architecture is shown in Fig.~\ref{img:hardware structure}. There are three main data and their corresponding buses, i.e., instructions, parameters and input images as well as two main modules, which are the calculation module, data selection module and data arrangement module.

There are two types of instructions: module execution instruction and parameter set instruction. When the module execution instruction is sent to the Instruction Bundle Table, the Instruction Bundle Table selects the low-level instruction bundle to be executed according to the correspondence between the model structure and the instruction bundle. Though relevant information (such as dimension and channel) set by the previous parameter set instructions, The Instruction Bundle Table calculates the number of executions, the number of data movements and storage addresses of each underlying module in the instruction bundle. Finally, it temporarily stores control signal in the register for controlling the sequential execution. Since the size of input image is 224, the efficiency will be constrained by bus transfer rate if all of them are taken out at once. Therefore, we design the data selection and data arrangement module to alleviate this phenomenon. The module selects and arranges the input images, and then transfer them to the data bus sequentially by means of parameter set instruction. In this way, computation and data transfer run in parallel. The parameters reach the computation module in stream form via the parameter flow bus. From the perspective of underlying computation, all parameters are used only once, so it is faster for parameter flow method.

The compute module is the core of the hardware architecture. It is divided into two categories according to characteristics of input data transfer, namely block module and stream module. There is positional relationship between the data of the underlying modules from block module, e.g., each pixel in input feature map has certain positional information. This correspondence leads to a heavy data dependency during computation and a certain random access. In contrast, the underlying modules from stream module participate in the computation on an element-by-element basis, with no positional relationships between all data and no computational dependencies in the computation. In other words, the data is retrieved in a sequential way during the computation, and there is no random access. So, it is possible to perform their computation in a stream form. 

\subsection{Storage structure}

Transformer-based visual models tend to transfer and compute the input data through a multidimensional tensor. There are two problems with such an approach. On the one hand, this description method makes the data of different scales and dimensions different structurally. Therefore, this description method has ambiguity in hardware implementation, because the priority between dimensions is not uniform in data selection, for example, when the operation of channel dimension is carried out, the model will describe the data by $[H, W, C]$ to simplify calculation. It is not uniform with the description of $[C, H, W]$, which is obviously is not consistent with the concept of universal design. On the other hand, these exists random read operations and transpose operations, which lead to incalculable latency. As a result, we design a new storage structure for describing data and parameters to solve the first problem. In the next section, we solve the second problem with data selection and data arrangement.
\begin{figure}[ht]
    \centering
    \includegraphics[width=\linewidth]{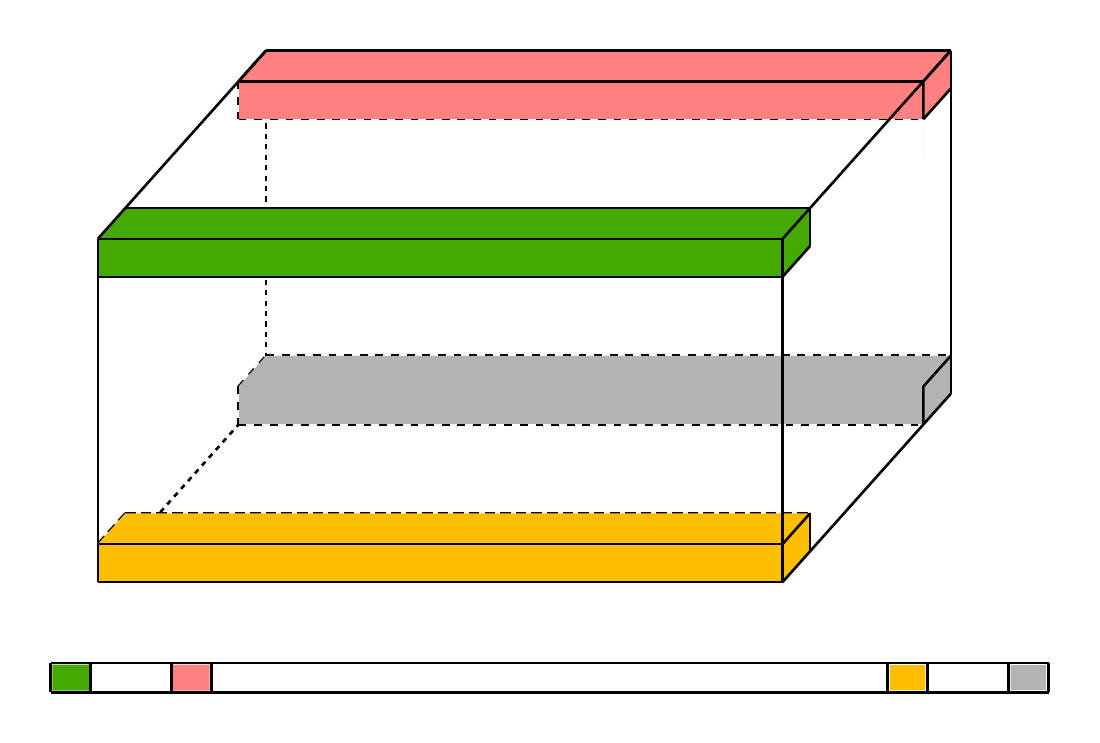}
    \caption{Access order based on our storage structure}
    \Description{Access order based on our storage structure}
    \label{img:storage structure}
\end{figure}
The whole idea of this new storage structure is to spread out the high-dimensional data to fit continuous storage characteristics of hardware storage. We take $C \times H \times W$ of data block at a time and map them sequentially into the hardware storage structure. Data selection module read the data along the dimension of $W$, and then in turn along the dimensions of $H$ and $C$, as shown in Fig.~\ref{img:storage structure}. This transformation of description method retains the characteristics of the software description method while solving its problems in hardware implementation. The new storage structure conforms to our idea of three-layer and two-level transformation structure, which does not change the data structure for the software level and does not cause additional Comprehension burden. For the hardware level, it is more in line with the characteristics of reconfigurable device storage structures.

\subsection{Data selection and data arrangement method}
This design performs data selection by passing 9 parameters to the data selection module, which are
\begin{itemize}
    \item The length, width and height of large cube ($MH$, $MW$, $MC$)
    \item The length, width, and height of small cube ($SH$, $SW$, $SC$)
    \item The offset of length, width and height of small cube ($FH$, $FW$, $FC$)
\end{itemize}
This form is also compatible with single or multiple channel and single or multiple dimension selection. With these parameters, this module can query the data blocks required by the subsequent modules. For more details, please refer to Algorithm.~\ref{algo:data-selection}.

\begin{algorithm}
    \caption{Data selection.}
    \KwData{base memory address $\operatorname{basePtr}$,  source address offset $src\_offset$, destination address offset $dst\_offset$}
    \For{$i \leftarrow 0$ \KwTo $SC-1$}{
        \For{$j \leftarrow 0$ \KwTo $SH-1$}{
            \For{$k \leftarrow 0$ \KwTo $SW-1$}{
                $mi \leftarrow i + FC$\;
                $mj \leftarrow j + FH$\;
                $mk \leftarrow k + FW$\;
                $\operatorname{basePtr}[i * SH * SW + j * SW + k + dst\_offset] = \operatorname{basePtr}[mi * MH * MW + mj * MW + mk + src\_offset]$\;
            }
        }
    }
    \label{algo:data-selection}
\end{algorithm}

\subsection{Computation module design}
\begin{figure}[ht]
    \centering
    \includegraphics[width=\linewidth]{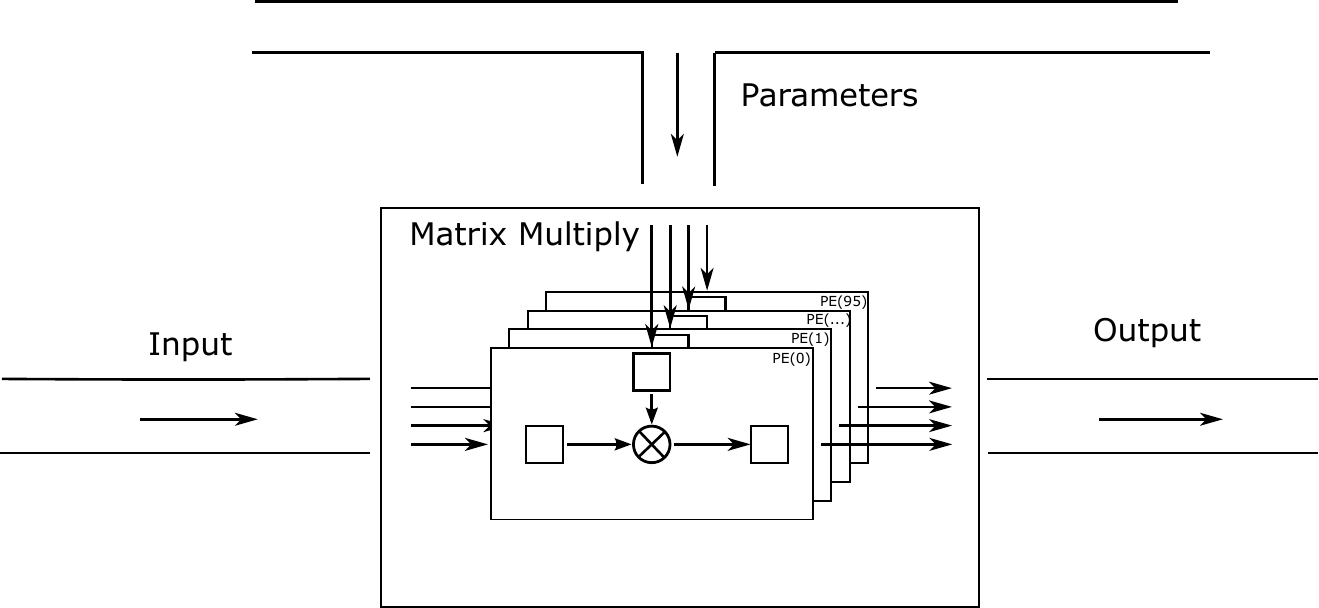}
    \caption{the hardware structure of stream module}
    \Description{the hardware structure of stream module}
    \label{img:stream module}
\end{figure}

Fig.~\ref{img:stream module} is a schematic diagram of the hardware implementation of stream module. The PE design of this module is simpler than block module because the input data in this module has no positional relationship and we can get input data and parameters directly according to the sequential storage structure mentioned in previous section. The input data enters the matrix multiply module and is assigned to C of PE units, where C is the size of Batch. Since the original parameters are multidimensional and not conducive to data access, we rearrange the parameters and transfer them sequentially to matrix multiply module though parameter flow bus. The parameters required for the computation are read as streams and assigned to PEs. Each PE cell get the corresponding C of parameters. Finally, the computation results are aggregated and output.

\section{Experiment result}
\subsection{Inference time}
During the experiment, we reproduced the official Swin Transformer tiny model that patch size is 4, window size is 7, and image size is $3 \times 224 \times 224$.
For the nonlinear functions, we implemented SoftMax by referring to the official implementation of pytorch; the Gelu function was implemented using the fitting function $0.5 \times (1 + \tanh(\sqrt{\frac{2}{\pi}}(x + 0.044715 \times x^3)))$ in \cite{hendrycks2016gaussian} to reduce computation stress; the variance calculation part of Layer Normalization function uses the formula $DX=E(X^2)-(EX)^2$ to circumvent the use of division as much as possible.

As shown in Table.~\ref{tab:inftime}, the experiments calculate the inference time and the proportion of each module when input is a $3 \times 224 \times 224$ image. Matrix Multiply module takes up the majority of inference time, as high as 70.6\%. This reflects the computation intensive nature of Transformer model, while the overall percentage of data selection and data arrangement module reaches 29.2\%, comparing this to show that our parameter flow design greatly reduces waiting time during data transfer. In the other three nonlinear functions, Gelu takes almost zero time, demonstrating the effectiveness of the Gelu function proposed in \cite{hendrycks2016gaussian}. The Layer Normalization function takes the longest time. Although we use a scheme that avoids the use of division as much as possible, the computation complexity of Layer Normalization is still on the large side, and there are still some cases where division is unavoidable. The final inference time for this experiment was $11.8ms$ for an image.

\begin{table*}
    \caption{Inference time of each module}
    \label{tab:inftime}
    \begin{tabular}{cccc}
        \toprule
        Module                              & Batch & Inference time(ms) & Percentage(\%) \\
        \midrule
        Matrix Multiply                     & 96    & 8.3492692          & 70.60          \\
        SoftMax                             & 96    & 0.00196            & 0.20           \\
        Layer Normalization                 & 96    & 0.01666            & 0.40           \\
        Gelu                                & -     & 0.00036            & 0.00           \\
        Data Selection and Data Arrangement & -     & 3.4572702          & 29.21          \\
        Pre-process                         & 96    & 0.00232            & 0.00           \\
        Total                               & -     & 11.8255194         & 100            \\
        \bottomrule
    \end{tabular}
\end{table*}

\subsection{Comparison with CPU, GPU, FPGA}

\begin{table*}
    \caption{Throughput (inference/sec) of vis-TOP compared with CPU, GPU, and FPGA. We also give relative throughput compared to FTRANS, throughput per DSP slice relative to FTRANS, NPE and FQ-BERT}
    \label{tab:compare}
    \begin{tabular}{ccccccc}
        \toprule
                            & CPU   & GPU          & \begin{tabular}{@{}c@{}}Ftrans\\ \cite{Li2020} \end{tabular} & \begin{tabular}{@{}c@{}}NPE(8-bit)\\ \cite{Khan2021} \end{tabular} & \begin{tabular}{@{}c@{}}FQ-BERT\\ \cite{liu2021hardware} \end{tabular} & Vis-TOP \\
        \midrule
        Throught            & 5.05  & 55.86        & 101.79                                                       & 135.14                                                             & 22.74                                                                  & 84.56   \\
        Relative Speedup    & 0.09x & 1x(baseline) & 1.82x                                                        & 2.42x                                                              & 0.41x                                                                  & 1.51x   \\
        DSP Slices Utilized & -     & -            & 6840                                                         & 2020                                                               & 1751                                                                   & 558     \\
        Throughput per DSP  & -     & -            & 0.0148                                                       & 0.0669                                                             & 0.0129                                                                 & 0.1515  \\
        \bottomrule
    \end{tabular}
\end{table*}

We have done comparison experiments on the GPU, CPU and FPGA respectively. In this experiment, the CPU is an Intel(R) Xeon(R) CPU E5-2695 v3 @ 2.30GHz and the GPU is a Tesla T4, both of whose results were obtained by running on Google's colab using the pytorch framework.

In comparison experiments with other works based on reconfigurable device, to our knowledge, no researcher has applied Transformer-based visual models to the field of reconfigurable computation, so this paper compares our design to other Transformer model in the NLP. FQ-BERT is a fully quantized Bert proposed by \cite{liu2021hardware}.  The 8-bit NPE of NVU-1024 \cite{Vaswani2017} is an overlay processor for BERT-based models. The FPGA-based FTRANS provides an example of an efficient Transformer model accelerator by running RoBERTa, which is an optimized version of well-trained BERT with the same model architecture. All are designed for BERT or its variants, and BERT has a similar underlying algorithm to Swin Transformer. Considering the size and structure of model, some metrics, such as latency, are difficult to compare horizontally, so the hardware measurement parameters chosen for this experiment are Throughput, DSP Slices Utilized and Throughput per DSP. Their comparison effect is not affected by the size and structure of models.

The experimental results are shown in Table.~\ref{tab:compare}. Compared to CPU and GPU, our throughput is 16x that of CPU and 1.5x that of GPU. This is due to the fact that we classify all operations of model and we implement the corresponding underlying module for each different type of operation. Thanks to that, our design achieves high parallelization and certain customization. Our Throughput is 3.71x that of FQ-BERT, but lower than that of FTRANS. Because that our design is three-layer and two-level transformation structure, compared to the customized accelerator, Vis-TOP divide the model in a more fine-grained way and the underlying modules are more universal. This feature allows our DSP Slices Utilized to be much lower than FTRANS. Due to the unique hardware and software co-design, our resource utilization is very high. As can be seen from the table, our Throughput per DSP is more than 10x that of non-overlay processor designs. This further illustrates the advantage of overlay processor in terms of resource utilization. In comparison to the NPE, which is also an overlay processor design, we demonstrate higher module reusability and our Throughput per DSP is still 2x that of the NPE without any special operation of non-linear functions. In a word, Vis-TOP achieves a balance between resource consumption and throughput.

\section{Conclusion and future work}

In this article, we propose Vis-TOP, an overlay processor based on software and hardware co-design on reconfigurable device. In addition to handling the deployment of existing Transformer-based visual models, Vis-TOP will not need to reconfigure the underlying hardware architecture but only modify instructions or add private module interfaces if a new Transformer-based visual models appears in the future. Vis-TOP has a three-layer and two-level transformation structure and a corresponding hardware architecture. This realizes the transformation and compatibility of software to hardware and software to hardware. Due to its structure, Vis-TOP has the programmability of software and the high parallelism of hardware and exist a large expansion space. At the same time, Vis-TOP can also meet the inference latency and resource consumption requirements in the image recognition environment. In general, vis-TOP is an effective solution for deploying Transformer-based visual models on reconfigurable devices.

In Vis-TOP, design of nonlinear functions such as SoftMax and Gelu is just customized module interfaces. In the future work, we will focus on non-linear functions. We will explore method of function fitting to reduce the calculation cost of nonlinear functions.

\newpage

\end{document}